\documentclass{article}

\usepackage{arxiv}

\usepackage[utf8]{inputenc} 
\usepackage[T1]{fontenc}    
\usepackage{hyperref}       
\usepackage{url}            
\usepackage{booktabs}       
\usepackage{amsfonts}       
\usepackage{nicefrac}       
\usepackage{microtype}      
\usepackage{lipsum}
\usepackage{graphicx}
\graphicspath{ {./images/} }

\usepackage{graphicx}%
\usepackage{amsmath,amssymb,amsfonts}%
\usepackage{amsthm}%
\usepackage{mathrsfs}%
\usepackage[title]{appendix}%
\usepackage{xcolor}%
\usepackage{textcomp}%
\usepackage{manyfoot}%
\usepackage{booktabs}%
\usepackage{algorithm}%
\usepackage{algorithmicx}%
\usepackage{algpseudocode}%
\usepackage{listings}%
\usepackage{multirow}%
\usepackage{caption}
\usepackage{subcaption}
\usepackage{tabularray}
\usepackage{hhline}
\usepackage{enumitem}
\usepackage{placeins}

\title{Continual Learning Approaches for Anomaly Detection}

\author{ {Davide Dalle Pezze} \\
	University of Padova\\
	\texttt{davide.dallepezze@unipd.it} \\
	\And
	{Eugenia Anello} \\
	University of Padova\\
	\texttt{anello@dei.unipd.it} \\
	\And
	   {Chiara Masiero} \\
	Statwolf Data Science Srl\\
	\texttt{chiara.masiero@statwolf.com} \\
 \And
	{Gian Antonio Susto} \\
	University of Padova\\
	\texttt{gianantonio.susto@unipd.it} \\
}

\begin{document}

\maketitle

\begin{abstract}
Anomaly Detection (AD) is a relevant problem in numerous real-world applications, especially when dealing with images.
However, in real-world applications, it is common that the input data distribution can change over time, decreasing performance significantly.
Therefore, in this study, we investigate the problem of Pixel-Level Anomaly Detection in the Continual Learning setting, where the model adapts to the new data while maintaining the knowledge of old data. 
We implement and test several AD techniques and adapt them to work in the CL setting using the Replay approach.
We evaluate them using the well-known MVTec AD Dataset, where each object corresponds to a new learning task.
Moreover, a significant challenge when dealing with the Replay approach is the memory occupied to store a portion of past images, which could be too heavy for many resource-constrained systems.
Therefore, we propose a novel approach called SCALE, which performs high compression levels while preserving image quality through Super-Resolution techniques.
Using the SCALE method to compress replay memory, in conjunction with the AD technique Inpaint, allows for obtaining the best AD results while significantly reducing memory consumption.
\end{abstract}



\section{Introduction}\label{sec1}

    Anomaly Detection (AD) is an important and challenging problem in the field of Machine Learning.
    Anomalies are patterns characterized by a noticeable deviation from the so-called normal data, where normal means compliant with some typical or expected features \cite{clvaead}.  
    In particular, when applied to the Computer Vision domain, AD finds many applications in the real world, such as in the industrial and medical domains. 
    For instance, AD techniques can detect defects in industrial products or irregularities in chest X-ray images.

    To address the problem of Anomaly Detection in the Computer Vision domain at the image-level and pixel-level, many of the proposed approaches in the literature are based on neural networks \cite{bergmann2019,zavrtanik2021reconstruction,yu2021fastflow}.
    An advantage of these approaches is that they are based on the unsupervised paradigm. This translates into easier adoption in many real-world applications since the label collection phase, which can be extremely expensive, especially for pixel-level annotations, can be avoided.

    Despite the recent advancements in the field, an issue persists.
    In real-world scenarios, shifts in the input data distribution can occur. 
    In these situations, the model must function successfully on older data while adjusting to the new input.
    For instance, new products could be added to the environment and we want to update the model to learn to detect the defects in the new products while detecting correctly the old ones.
  
      However, the neural networks are prone to Catastrophic Forgetting (CF). When learning a new task, artificial neural networks forget the previous ones \cite{vandevan2019}.
      This aspect significantly hinders the adoption of these approaches in real-world scenarios. 
    A new branch of Machine Learning has been introduced, known as Continual Learning (CL), to address this issue \cite{de2021continual}. CL focuses on learning from a stream of tasks to adapt to the new incoming tasks while remembering the previous ones. 

In this work, we study the framework of Continual Learning for the Anomaly Detection problem (CLAD), with a focus on Anomaly Localization, also known as Pixel-Level Anomaly Detection.
We implement several well-known approaches for AD in the Computer Vision domain and test their performance in the CLAD setting. 

In particular, we adapt these AD techniques to work in the CL setting by employing a well-known CL method called Replay.
Replay is the most well-known and effective method for preventing forgetting \cite{Replaypaper,Replaymethods} in the supervised image classification problem.
However, whether it will still be effective for the Anomaly Detection problem cannot be known a priori and needs to be investigated.
Therefore, as the first approach to be tested in the CLAD setting, we consider the Replay method.
This method stores a portion of the old data in an additional memory. While training on a new task, the new data is combined with the data from the memory to maintain knowledge of old tasks.

One of the main issues with Replay is the requirement for additional memory, which is called replay memory.
This can be a strong constraint, especially for resource-constraint devices that need to work with small memory and processing power.
To address this issue, we consider Compressed Replay approaches to reduce the replay memory size by compressing the stored samples.
Moreover, we propose a novel technique, SCALE, which applies a Super Resolution model to compress the images in the replay memory.

\begin{figure*}  
        \centering
        \includegraphics[width=1\textwidth]{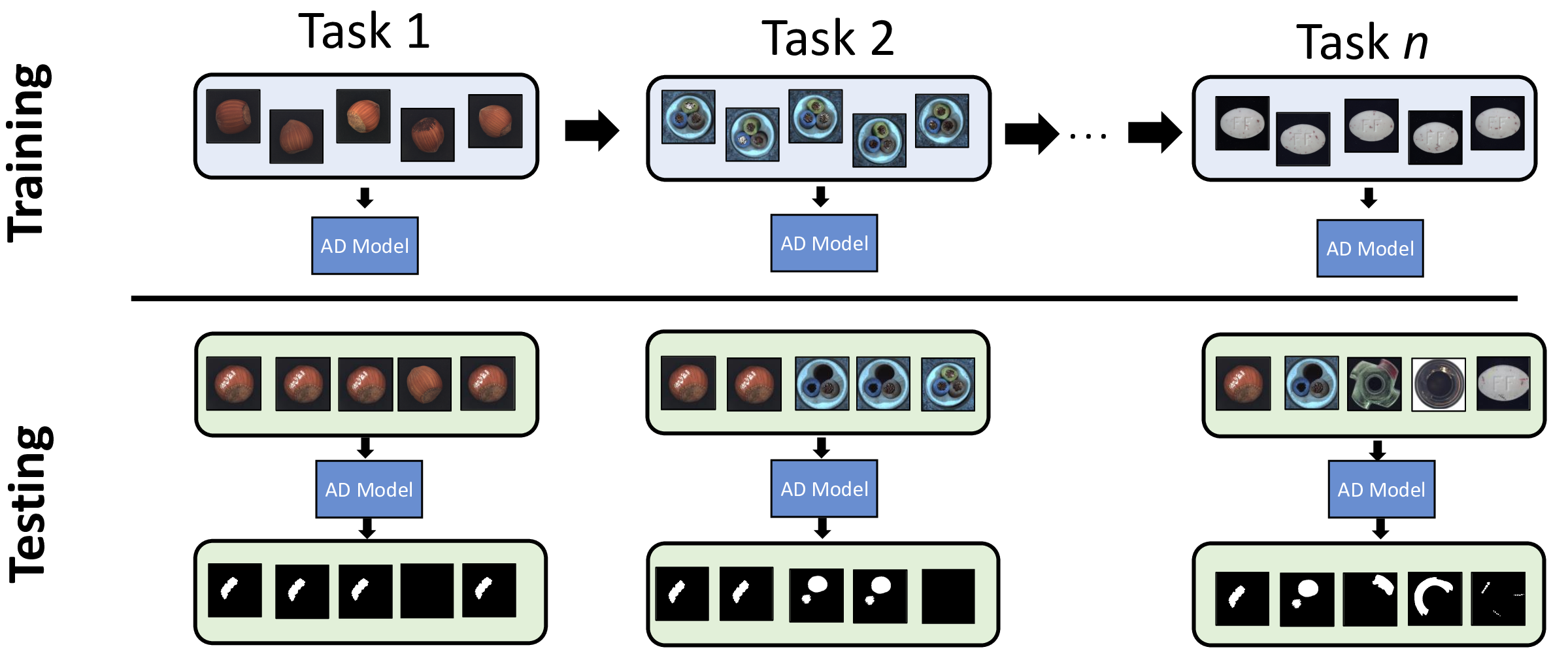}
        \caption{Scheme of the Continual learning for Anomaly Detection (CLAD). Over time, new items (tasks) arrive. The model must learn the new items and differentiate between anomalous and normal samples. In the testing phase, each pixel is classified as normal (black) or anomalous (white). During this phase, the model is evaluated on both new products and past ones to ensure that previous knowledge is maintained.  }
         \label{Fig:Scenario_CLAD}
    \end{figure*}

Using the MVTec dataset, we demonstrate Replay's efficacy in the CLAD setting and SCALE's optimal compression.
The MVTec dataset is a well-known dataset for studying AD at image-level and pixel-level composed of 10 industrial products \cite{bergmann2019}.
Therefore, we introduce a CL benchmark based on the MVTec dataset as a stream of tasks by considering each object as a distinct task, as represented in the scheme of Figure \ref{Fig:Scenario_CLAD}.
In summary, the key contributions of this work are as follows:
\begin{itemize} 
\item We study the novel problem of Pixel-Level Anomaly Detection in the Continual Learning setting.
\item We implement several AD techniques and adapt them to work in the Continual Learning.
\item We propose a novel technique called SCALE to compress the replay memory to allow the execution on resource-constrained systems.
\end{itemize}

The outline of the paper is as follows:
In Section \ref{Sec:RelatedWorks}, we describe related work in terms Continual Learning, Anomaly Detection and works about AD in the CL setting.
In Section \ref{Sec:ProposedFramework_ADCL}, we introduce the framework proposed for AD and describe the CL methods and AD architectures considered.
In Section \ref{Sec:SCALE}, we describe our novel approach for Compressed Replay called SCALE based on an SR model to retain the knowledge of old tasks.
In Section \ref{Sec:Experimental Settings}, we describe the experimental setup that implements our approach, including details about the evaluation metrics used, the dataset, and the CL setting proposed.
Finally, in Section \ref{Sec:Results}, we present our findings before concluding in Section \ref{Sec:Conclusion}.

\begin{table}[th]
\centering
\begin{tabular}{|c|c|}
\hline
\multicolumn{1}{|l|}{\textbf{Notation}} & \textbf{Description}                  \\ \hline
AD                                      & Anomaly Detection                     \\ \hline
CV                                      & Computer Vision                      \\ \hline
CLAD                                    & Continual Learning for Anomaly Detection                    \\ \hline
CF                                      & Catrastrophic Forgetting              \\ \hline
CL                                      & Continual Learning                    \\ \hline
DL                                      & Deep Learning                         \\ \hline
SR                                      & Super Resolution                       \\ \hline
\end{tabular}
\caption{This table presents a concise reference for the notations employed in the course of our research. These notations will be used extensively in our discussions and analyses.
}
\end{table}

\section{Related Work}
\label{Sec:RelatedWorks}
In the following text, we describe the concepts of Continual Learning in Sec. \ref{subsec:RelatedWork_CL}. Then, Sec. \ref{subsec:RelatedWork_AD} describes the literature on Anomaly Detection for the Computer Vision domain.
Finally, we discuss previous work for Anomaly Detection in the Continual Learning scenario in Sec. \ref{subsec:RelatedWork_CLAD}.

\subsection{Continual Learning}
\label{subsec:RelatedWork_CL}
Continual learning allows machine learning models to update and expand their knowledge over time, with minimal computation and memory overhead \cite{de2021continual}.
Thus, an effective CL solution is expected to have low forgetting, require low memory consumption, and be computationally efficient.
CL methods are usually classified into three categories: \textit{replay-based methods} \cite{shin2017continual,rolnick2019experience,chen2023online,atkinson2021pseudo,luo2023relay}, \textit{regularization-based methods} \cite{pomponi2020efficient, ramapuram2020lifelong,han2022online} and \textit{architecture-based methods} \cite{sokar2021spacenet, mallya2018piggyback,rajasegaran2019random}.

The related literature suggests that the replay-based approaches are the most effective and practical solution to reduce forgetting \cite{Yangemp,pellegrini2020latent, buzzega2021rethinking,  kim2020imbalanced}.
Among replay-based approaches, a very well-known one is Experience Replay (ER) \cite{rolnick2019experience, chaudhry2019tiny}. It consists of storing some randomly selected samples from the previous tasks in the raw format and using them to maintain knowledge about previous tasks when training a new task. 

It is crucial to note that this approach needs additional space to memorize the subset of images from previous tasks.
This implies that memory storage capacity is a major limitation; the model's performance decreases as memory capacity decreases.
A possible solution comes from the approach \textit{Generative Replay}. The idea consists of having a generative model like a GAN or VAE to generate images (even separated by the main model). 
This can be very useful because it eliminates the need to save any data in memory. 
This method proved effective in previous experiments when learning simple distributions like MNIST.
However, when learning complex distributions like CIFAR-10, Generative Replay exhibits a high level of forgetting, making it challenging to consider it a viable solution in real-world scenarios \cite{lesort2019generative}.

Recently, some methods that we denote with the generic term of \textit{Compressed Replay} try to reduce the cost of storing old samples in the replay memory by compressing the old data representation.
With less memory required for each sample, more samples can be stored in the same memory space, enhancing the model’s performance.

For instance, \cite{pellegrini2020latent} proposes a novel method called Latent Replay. As the name suggests, they store activations (less costly than the images) from an intermediate neural network layer and store them instead of the images. Then, replay is applied only on the last layers using these compressed representations.
Similarly, in \cite{hayes2020remind}, the lower layers are frozen, and only the upper layers are trained. Moreover, to store more samples, they propose using product quantization (PQ) to compress and save the features efficiently, obtaining a high compression factor. In \cite{iscen2020memory}, authors propose to apply feature adaptation so that features learned for the old task remain consistent with the new feature space.

While these works show interesting results, the mentioned approaches work only for the supervised classification problem.
In addition, most Unsupervised AD approaches require working with raw data. In other words, we are more interested in the generative problem than the classification one.
Therefore, we want to compress the original samples while retaining the ability to reconstruct the original image in the output, which is not possible with the methods described above. 

To overcome such a challenge, we propose a novel method called \textit{SCALE}, described in Sec. \ref{Sec:SCALE} and demonstrating in Sec. \ref{subsec:Results_FID} it can achieve a high level of compression with a high level of quality in the generated images.

\subsection{Approaches for Unsupervised Anomaly Detection in the Computer Vision domain}
\label{subsec:RelatedWork_AD}

Unsupervised Anomaly Detection approaches can be used for many Computer Vision problems, such as those in manufacturing, the medical field, autonomous vehicles, and surveillance systems.
In these domains, detecting anomalies is critical for safety, security, and efficiency.

In addition, having these systems interpretable can help integrate them better into the flow of a realistic application.
This is achieved by the capability of the AD techniques to extend the model's predictions beyond image-level and into pixel-level granularity.
Therefore, these AD models allow the end user to determine if the item is anomalous and identify the specific parts of the image considered abnormal by the model.

Furthermore, since the entire process is unsupervised, gathering labels at the pixel and image levels is unnecessary, which can be very expensive, particularly for pixel-level annotations.

An example of an application is Quality Prediction in a manufacturing plant. A model analyzes items on a conveyor belt, identifies the defective ones, and sends a warning to the manufacturer. Then, in line with Industry 5.0, which must be user-centric, the manufacturer can understand why this item is considered anomalous by looking at the pixel-level anomalies that represent the defective parts. In other words, the model became interpretable with the ability to explain the predictions of anomalous items using pixel-level granularity.

Another instance is the medical field, in the analysis of chest X-ray, Brain MRI, Liver OCT, and Retinal OCT \cite{bao2023bmad}. In these cases, identifying abnormal chest X-ray images is not enough; the model must be able to interpret why that image is considered abnormal. Therefore, to aid medical professionals, it is essential not only to identify possible abnormal patients but also the reason for this prediction, and in such a context, the identified abnormal pixels could be helpful. 

While the AD techniques have interesting properties like interpretability and avoiding the label collection phase, they lack scalability.
Continual learning approaches can address this limitation by incrementally learning from new products as they become available.

We can split most of the Pixel-Level Anomaly Detection approaches into two main families: \textit{reconstruction-based methods} and \textit{feature embedding-based methods}.

\textbf{Reconstruction-based methods} learn to reconstruct normal images during training. However, images with defects are not expected to be well reconstructed during the evaluation. Consequently, they should exhibit a higher reconstruction error than normal images. Reconstruction-based methods include neural network architectures like autoencoders  \cite{bergmann2019,colinstain,ZAVRTANIK2021107706}, variational autoencoders  \cite{vaerec} and generative adversarial networks \cite{akcay2018ganomaly, SchleglSWSL17}. 
The idea is that autoencoders will ignore minor details that don't belong to the training distribution and reconstruct the image without defects.
However, they can sometimes reconstruct defects in output, leading to a failure to distinguish between abnormal and normal images.
In \cite{zavrtanik2021reconstruction}, an approach based on inpainting is proposed called RIAD. The idea is to hide a portion of the input image and train the model to reconstruct it.

\textbf{Embedding similarity-based methods}, on the other hand, use pre-trained neural networks to extract meaningful features from images.
These methods prove to have optimal results like PatchCore \cite{roth2022towards} and SPADE \cite{spade2021}, which use frozen feature extractors to save features in a so-called memory bank which is used in inference to detect the anomalous regions.
However, a significant limitation of these methods is the absence of learnable weights, which makes performing Continual Learning on such approaches challenging and not straightforward.
Other works like FastFlow \cite{yu2021fastflow} and CFLOW-AD \cite{gudovskiy2022cflow} use Normalizing Flow (NF) models \cite{rezende2015variational,kingma2018glow} to model the distribution of image features into a normal distribution.
Then, it becomes possible to use the probability of the data under the distribution to measure its normality.

\subsection{Continual Learning techniques for Anomaly Detection}
\label{subsec:RelatedWork_CLAD}
Even though AD is highly relevant in practice, only a few works in the literature address the problem of Continual Learning for Anomaly Detection.
Most of the existing works consider tabular data or time series.
In AD for Manufacturing, authors of \cite{maschler2021regularization} propose a regularization-based approach and evaluate it on a real industrial metal forming dataset.
For the domain of Financial Accounting, \cite{hemati2021continual} aims to learn from a stream of journal entry data experiences. The goal is the detection of fraud in non-stationary journal entry distributions. 
In the domain of Network Intrusion Detection, a series of works have been proposed in the literature.
In \cite{clvaead}, the authors propose using the Variational Autoencoder (VAE) with the Generative Replay approach to learn new items while performing AD continually. The datasets used in the evaluation are KDD Cup 1999 and MNIST, which have been artificially adapted to AD by marking one of the classes as anomalous. In \cite{amalapuram2022continual}, using the same dataset KDD Cup '99 and other datasets belonging to Network Intrusion Detection, the authors study the problem of AD formulated as a binary classification supervised learning. 
With the same spirit, another work in the field was conducted by \cite{gonzalez2022steps} using Generative Replay with a Dilated Convolutional Neural Network and validated on multivariate time-series datasets.

As for Anomaly Detection in Computer Vision, few works have been conducted.
\cite{doshi2022rethinking} presented a novel technique utilizing replay in surveillance videos (video-based anomaly detection). Despite the intriguing nature of the approach, the AD methodology introduced significantly diverged from the state-of-the-art techniques established for image-based anomaly detection.
A study considering a sequence of tasks for AD was performed in \cite{frikha2021arcade}. In contrast to supervised and unsupervised paradigms, the authors propose a meta-learning approach. 
Nevertheless, it is crucial to note that their methodology exclusively operates at the image-level and involves treating an entire class as anomalous, resulting in a scenario that diverges significantly from realistic situations.
The work presented in \cite{gabbar2023incremental} introduces a Continual Learning approach for supervised anomaly detection at the image-level, specifically in X-ray computed tomography images. 
Their investigation focuses on the automated tool inspection for a nuclear power plant.
The proposed solution addresses this challenge using regularization-based approaches.
While they study and solve a realistic problem, their study exclusively examines image-level AD through supervised classification.
This can limit use cases, especially when an unlabeled dataset or interpretability is essential.
On the contrary, \cite{li2022towards} advocates for an Unsupervised AD technique that retains samples from previous tasks. Despite moving from the supervised paradigm to the unsupervised one, this method exclusively operates at the image-level, diminishing its practicality in real-world scenarios that demand interpretability.
A comprehensive exploration of various AD techniques was performed in \cite{xie2023iad} by assessing multiple paradigms, including the CL setting. 
The study compared several AD methods, demonstrating how employing these methods using the Fine-Tuning approach (without CL solutions) leads to forgetting. Notably, the study does not present solutions for adapting these methods to function effectively within the CL setting.

Most of the work in computer vision focuses on predicting whether an image is normal or abnormal. However, Pixel-Level Anomaly Detection is frequently required in practice, given its several advantages, such as interpretability. Not only is it necessary to explain the predictions, but it is also necessary to perform root cause analysis.
Consequently, we examine the performance of established pixel-level AD techniques, assessing that the Fine-Tuning approach performs poorly in a stream scenario since it forgets previously seen data.
Subsequently, we introduce an adaptation to enable their effective operation in Continual Learning (replay-based approaches).
These methods are then evaluated on a complex image dataset such as MVTec, which contains pixel-based anomalies.

\FloatBarrier
\section{Proposed CLAD Framework}
\label{Sec:ProposedFramework_ADCL}

\subsection{Problem Definition}
\label{subsec:problem_definition}
As discussed, Unsupervised Anomaly Detection for the Computer Vision domain is a relevant problem in many real-world applications with several advantages, such as avoiding the costly label collection phase (unsupervised learning) and providing interpretability to the user by providing the specific pixel that causes to asses an image as abnormal.

Even though these works obtained excellent results, getting close to optimal performance, each new item typically uses a different model, which raises the issue of scalability.
Moreover, even if we consider training a single model with all different items, it is unrealistic to assume that in a real environment, all data are available from the beginning, and there won't be data distribution shifts over time.
Therefore, we consider a more realistic framework, Continual Learning for Anomaly Detection (CLAD).

The first objective of this study is to investigate how AD approaches perform when learning multiple items from the same model sequentially.
We then define a framework for CLAD that can be used for multiple CL approaches.
Our framework is proposed to be integrated into three CL approaches: Replay (Fig. \ref{fig:CLAD_framework_scheme_Replay}), Compressed Replay approaches (Fig. \ref{fig:CLAD_framework_scheme_CompressedReplay}), and Generative Replay \ref{fig:CLAD_framework_scheme_GenerativeReplay}.

The goal is to train a neural network model that can detect anomalies while retaining the knowledge it learned from previous tasks. A representation of this scheme is shown in \ref{Fig:Scenario_CLAD}. The model is trained on a sequence of tasks, each corresponding to one or more items. During training, the model learns the distribution of normal (i.e., anomaly-free) images for each task. Then, during evaluation, the model is tested on normal and abnormal samples of all seen tasks. This allows the model to detect anomalies at the pixel-level in the test data by comparing it with the learned distribution of normal data.

In more formal terms, in the CLAD setting, the model is trained on a sequence of tasks with a total of $T$ tasks. Each Anomaly Detection task $t$ corresponds to a dataset $D_t$.
Each dataset $D_t$ consists of pairs of data $(X_t, Y_t)$, where $X_t$ is a set of images or more formally $X_t \subset \mathbb{N}^{H\times W\times 3}$ where $H$ and $W$ denote the spatial dimensions and $t = 1 \ldots T$.
$Y_t \subset \{0,1\}^{H\times W}$ is a set of labels for each pixel of the image, indicating whether the pixel is normal (0) or anomalous (1).
The goal of the model is to learn a mapping $f_{\theta}: X \rightarrow [0,1]^{H\times W}$ from the space of images $X$ to a probability space,  assigning a probability to each pixel in an image, indicating how likely it is to be normal or anomalous.

\subsection{CLAD Framework}
\label{subsec:clad_framework}

\begin{figure}[!th]
    \centering
    \begin{subfigure}[t]{0.75\textwidth}
        \centering
        \fbox{
        \includegraphics[width=\textwidth]{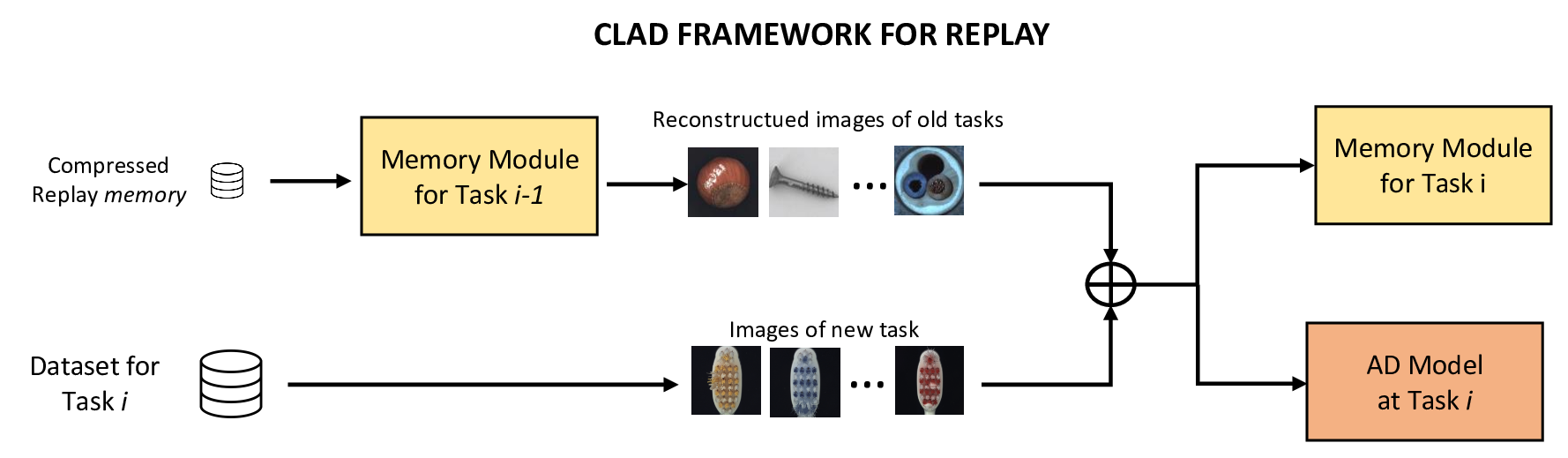}
        }
        \caption{CLAD Framework for Replay}
         \label{fig:CLAD_framework_scheme_Replay}
    \end{subfigure}
    \begin{subfigure}[t]{0.75\textwidth}
        \centering
        \fbox{
        \includegraphics[width=\textwidth]{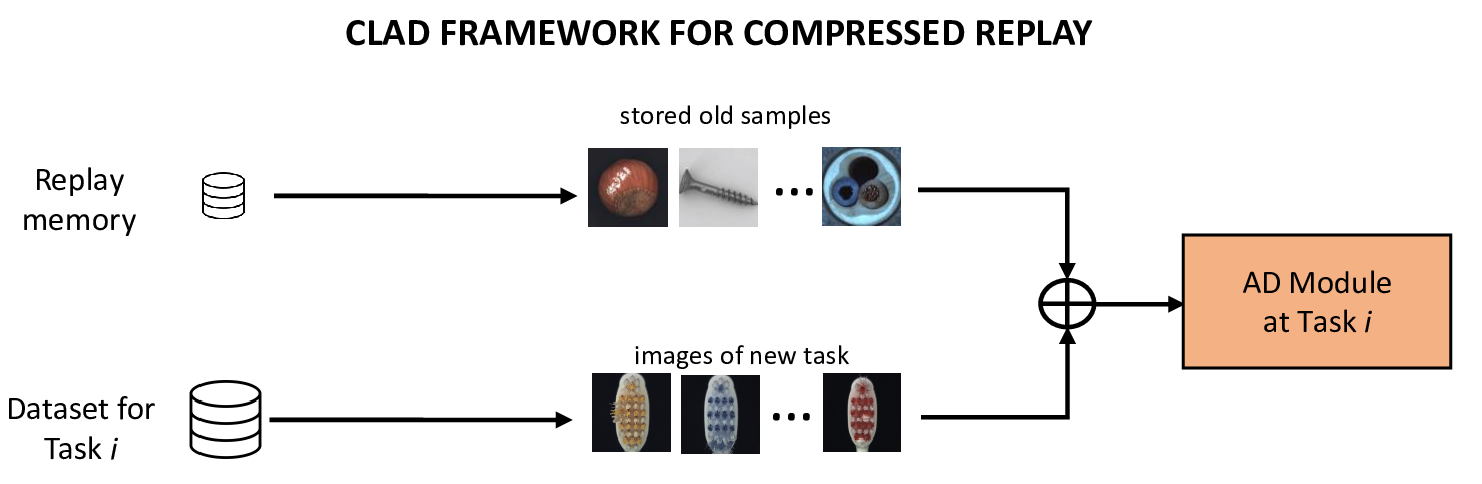}
        }
        \caption{CLAD Framework for Compressed Replay}
         \label{fig:CLAD_framework_scheme_CompressedReplay}
    \end{subfigure}
    \begin{subfigure}[t]{0.75\textwidth}
        \centering
        \fbox{
        \includegraphics[width=\textwidth]{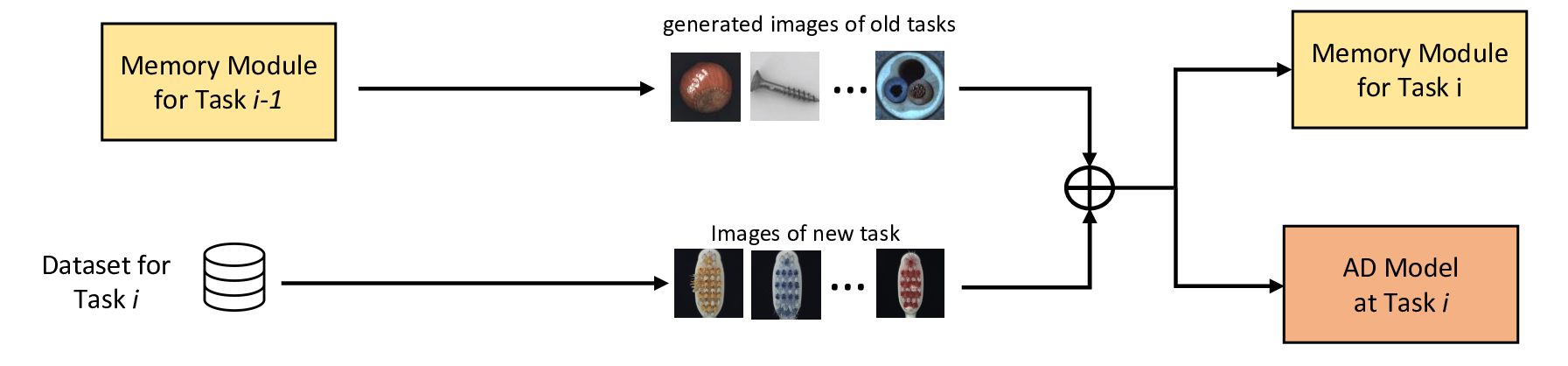}
        }
        \caption{CLAD Framework for Generative Replay}
         \label{fig:CLAD_framework_scheme_GenerativeReplay}
    \end{subfigure}

    \caption{
    (a) is the proposed CLAD framework for the Replay. Here, the Memory Module is represented by the replay memory that contains past samples. 
    (b) is the proposed CLAD framework for the Compressed Replay. In this case, the Memory Module (a model) is used to reconstruct images from the compressed samples of old tasks. Note that it is accepted (but not mandatory) that a single model can act simultaneously as a Memory Module and an AD Module (e.g., the VAE).
    Then, (c) is the CLAD framework for the Generative Replay. Here, the Memory Module is a generative model that creates images belonging to old tasks as needed.
 }
    \label{Fig:CLAD_framework_scheme}
\end{figure}

In this study, we introduce a comprehensive framework designed to address the challenges of CLAD. The primary objective of this framework is to establish a versatile and adaptable structure that accommodates existing methodologies explored in this study while remaining open to future integrations of other approaches. 
In particular, we propose a scheme for Replay depicted in Fig. \ref{fig:CLAD_framework_scheme_Replay}, a scheme for Compressed Replay in Fig. \ref{fig:CLAD_framework_scheme_CompressedReplay}, and a scheme for Generative Replay in Fig. \ref{fig:CLAD_framework_scheme_GenerativeReplay}.
Moreover, with some modifications, our framework easily adapts to other CL approaches, such as EWC \cite{kirkpatrick2017overcoming} and LwF \cite{li2017learning}.

The proposed framework, depicted in Fig. \ref{Fig:CLAD_framework_scheme}, is broken down into two modules:

\begin{enumerate}
    \item \textbf{Memory Module}: The Memory Module is a key component of the proposed CLAD framework, \textit{responsible for storing and updating information about previous tasks}. Depending on the type of CL method considered, such a module may take various forms. In the case of Replay, the Memory Module simply stores a limited number of images from previous tasks (see Fig. \ref{fig:CLAD_framework_scheme_Replay}). 
    In the case of Compressed Replay, the Memory Module is represented by a model, like the autoencoder (AE) (see Fig. \ref{fig:CLAD_framework_scheme_CompressedReplay}), which, in conjunction with a compressed memory, produces samples of the old tasks to combine with the new data.
    It is also possible to use Generative Replay with VAE and GAN models; in this case, the Memory Module consists of only a generative model used to produce images of old tasks to concatenate with the images of the current task (see Fig. \ref{fig:CLAD_framework_scheme_GenerativeReplay}).

    \item \textbf{AD Module}: \textit{It is the architecture used to detect anomalous pixels.} The chosen architecture can be any of the previously described AD approaches in Sec. \ref{subsec:RelatedWork_AD}, such as CAE, VAE, SR, and Inpaint. When performing Replay or Compressed Replay, the Memory Module retrieves the images of previous tasks and concatenates them with the data of the new task to update the AD Module, as depicted in Fig. \ref{Fig:CLAD_framework_scheme}.    
    \end{enumerate}
    
\textbf{When a model can act as both Memory and AD modules}
It is crucial to highlight that while the Memory Module and the AD Module are commonly discussed as distinct functional components, it is also possible that a single model can fulfill both roles. 
For instance, various reconstruction-based techniques, like the VAE, can generate images related to previous tasks. 
Yet, this reconstructive capability can also be used to perform Anomaly Detection by comparing the disparity between the reconstructed and original images (previously described as reconstruction-based approaches).
In some cases, the same model can function concurrently as both a Memory Module and an AD Module, whereas in other scenarios, separate models are required for Memory and Anomaly Detection.
For instance, the SR model could be used just as a Memory Module, while an AD approach like Inpaint could be used to act as an AD Module.

\FloatBarrier
\subsection{Considered Benchmark Dataset}
\begin{figure*}[th]
    \centering\includegraphics[width=1.0\textwidth]{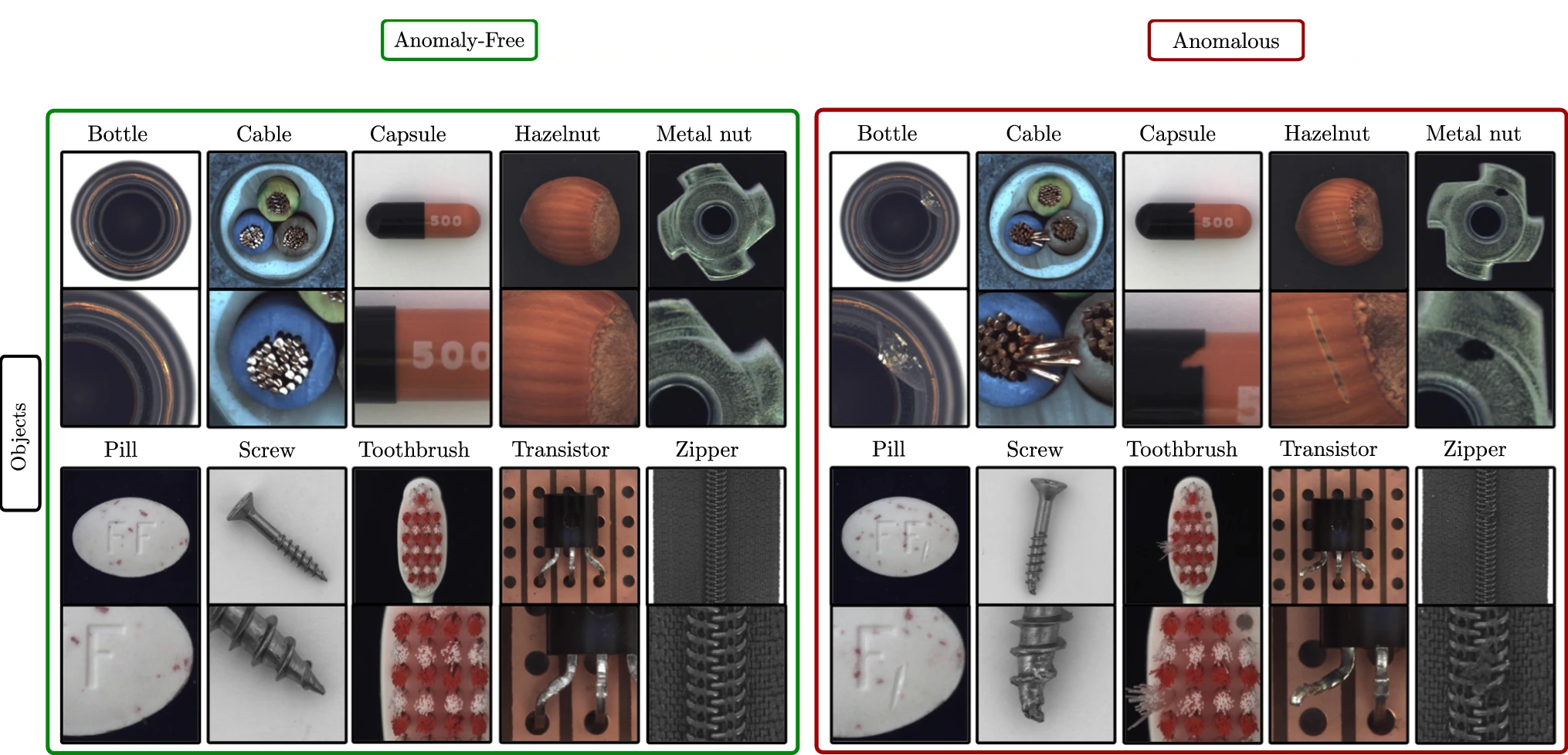}
  \caption{Image examples from the MVTec Dataset AD. For each object is shown a normal sample (in green) and an anomalous sample (in red). For each sample, the entire object is displayed, as well as a zoomed-in view of the region containing the defect. The samples illustrate the range of variations and abnormalities found in the dataset.}
  \label{Fig:MVTec Dataset AD}
\end{figure*}

We consider the MVTec Dataset \cite{bergmann2019} as a benchmark to study the performance of architectures and CL methods for Anomaly Detection.
The MVTec AD dataset is a recent and extensive industrial dataset that includes 5354 high-resolution images divided into 15 categories: 5 types of textures and ten types of objects (see Fig. \ref{Fig:MVTec Dataset AD}). The training set only contains normal images, whereas the test set includes defect and defect-free images. 
The image resolutions range from (700,700,3) to (1024,1024,3) as resolution.

In our CLAD framework, we will consider a sequence of 10 tasks, each corresponding to one of the ten objects.
In this study, the model is presented with a series of tasks, and the abnormal pixels for each object must be identified, as depicted in Fig. \ref{Fig:Scenario_CLAD}. 
The MVTec Dataset used is more challenging and complex than the commonly used datasets MNIST and CIFAR-10, which are often used in Continual Learning literature \cite{chen2018lifelong, diaz2018don, lopez2017gradient}. 
These datasets have smaller image sizes (28,28) for MNIST and (32,32) for CIFAR-10 and fewer images, making them less representative of real-world scenarios than the MVTec dataset.

\section{SCALE: A novel approach for Compressed Replay}
\label{Sec:SCALE}
\label{Sec:our_approach}
Even if Replay is proven effective in the CLAD setting, its application in real-world scenarios could be unfeasible due to memory limitations.
For this reason, we consider some Compressed Replay techniques based on CAE and VAE models, which aim to reduce the memory replay footprint.

Moreover, we introduce a novel Compressed Replay method called SCALE (SCALing is Enough),  based on the  Super Resolution problem.
Super Resolution (SR) aims to convert a low-resolution image with coarse details into a high-resolution image with improved visual quality and refined details \cite{anwar2020deep}.
In practice, as an SR model, we consider the Pix2Pix \cite{isola2017image}, a general-purpose model for image-to-image translation composed of a conditional GAN (cGAN) \cite{mirza2014conditional}. 
Though such architecture is not state-of-the-art in the SR field, it is sufficient to obtain good results and justify its use in this context, as shown in Section \ref{Sec:Results}. 
    Therefore, in our work, the SR model is studied with three final objectives:
    \begin{itemize}
        \item Since it is the first time that the SR problem is studied in CL, we provide an evaluation in terms of the reconstruction of the original images and how this approach behaves in terms of forgetting. 
        \item We study its efficacy as memory to retain the knowledge of old tasks. We evaluate the model in terms of compression factor and quality of the reconstructed images. 
        \item We also investigate the use of memory in combination with AD approaches that are, by nature, unable to perform the Memory Module function, like the Inpaint technique.
    \end{itemize}

\begin{figure*}[th]   
        \includegraphics[width=1\textwidth]{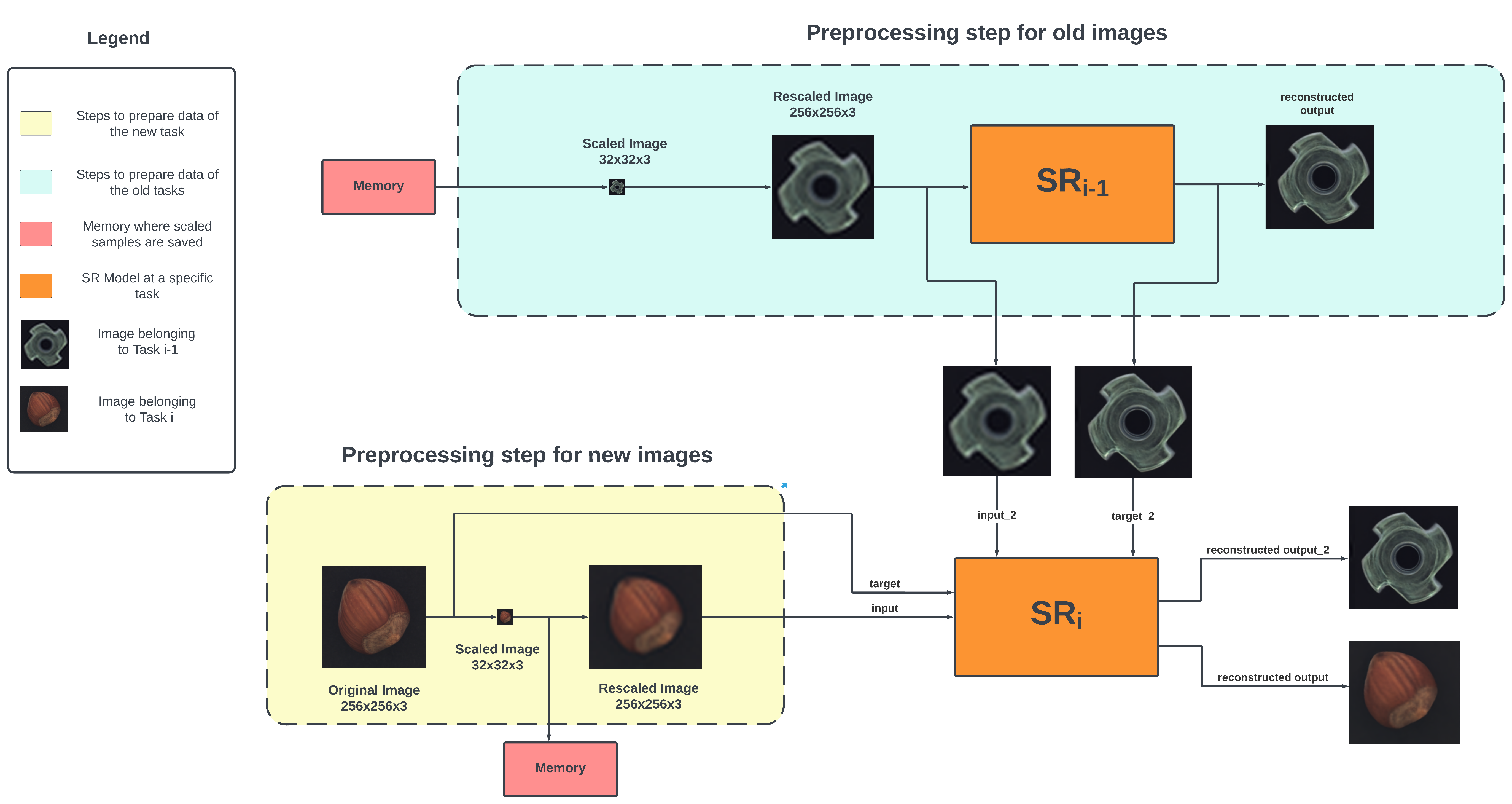}
        \caption{Scheme of the proposed approach for compressed Replay called SCALE. $SR_i$ represents the Super Resolution model at task $i$. First, the current task images are scaled (compressed) and saved in memory (yellow area). Then, they are rescaled ($input$), and the model is trained to reconstruct the original image ($target$). Instead, for old tasks, the images are taken from memory, rescaled, and reconstructed using the SR model from the previous task to retain old knowledge (light blue area). After that, the rescaled blurry image ($input_2$) is fed into the current SR model, with the target corresponding to the image reconstructed by SR in Task $i-1$ (i.e., $target_2$). }
         \label{Fig:framework_SR}
    \end{figure*}

The proposed schema for using the SR model in Continual Learning for Compressed Replay is shown in Fig. \ref{Fig:framework_SR} and can be summarized in the following steps:
\begin{enumerate}
    \item When a new task is received, some images are randomly selected, like in the Replay. However, contrary to Replay, the images are resized to a lower resolution before being stored in the replay memory, reducing the memory footprint.  This part is shown in the yellow portion of Fig. \ref{Fig:framework_SR}, the image is scaled from 256x256x3 to 32x32x3 and saved in the Memory (red rectangle of Fig. \ref{Fig:framework_SR}).   
    
    \item To retrieve the images of previous tasks, saved images are taken from the memory in low-resolution (32x32x3) and resized in the original size (256x256x3), still in low-resolution. Then, the images are fed to the SR model to obtain high-resolution output images. This process to retrieve images of old tasks is depicted in the light blue portion of Fig. \ref{Fig:framework_SR}.   
    
    \item Utilizing the images of the current task (depicted in the yellow segment of Fig. \ref{Fig:framework_SR}) alongside the reconstructed images from previous tasks (as illustrated in the light blue portion of Fig. \ref{Fig:framework_SR}), we concatenate these images and then supply the entire batch as input to the network for the update. 
\end{enumerate}

Therefore, by memorizing the scaled images and increasing the resolution when needed, a high level of compression can be achieved in the replay memory.
It should also be noted that the value of such compression depends on the desired final quality, and it will depend on the final scope. 
We are going to present the results of SCALE in terms of performance for Anomaly Detection in Section \ref{subsec:comparison_AD_SCALE} and in terms of the quality of reconstructed images in Sec. \ref{subsec:Results_FID}.

\FloatBarrier

\section{Experimental Settings}
\label{Sec:Experimental Settings}
\label{sec:experimental_setting}

We provide all the details of the experiments below.
In Sec. \ref{subsec:experimental_setting_models} all the information on the tested AD models are provided.
Then, in Sec. \ref{subsec:considered_cl_strategies} are described the different CL techniques tested in the experiments.
Eventually, we describe the metrics considered to evaluate AD, image reconstruction quality, and for the CL setting in Sec. \ref{subsec:metrics_AD}, \ref{subsec:metrics_image_quality}, and \ref{subsec:metrics_cl}, respectively.

\subsection{Considered Architectures in Memory and Anomaly Detection modules}
\label{subsec:experimental_setting_models}
The following models were considered as both Memory and AD modules:
\begin{itemize}
    \item \textbf{Convolutional Auto-Encoder(CAE)}: It has a latent space of shape (512,4,4), i.e., of dimension 8192, implying a compression factor value equal to 6 (assuming that we are working with 4 bytes for each value). Taking into account an input shape (256,256,3) we can calculate the compression factor as $\frac{196608}{8192 \cdot 4}=6$. We will memorize the activations of the CAE obtained in the latent space in the case of Latent Replay.
    
    \item \textbf{Variational Auto-Encoder(VAE)}: Using the VAE (Variational Autoencoder) architecture, we were able to achieve good results with a smaller latent space compared to the CAE (Convolutional Autoencoder). Specifically, we obtained a latent space dimension of 256 and a compression factor of 196. 

    As pointed out before, this model can be used to perform Compressed Replay as for the CAE, but also as Generative Replay since it can generate images without further visual inputs.
    The quality of VAE to compress and memorize compressed images to act as a Memory Module in the Compressed Replay is discussed in Sec. \ref{subsec:Results_FID}.
    Moreover, since VAE can also be used for AD, the performance for Replay is discussed in Sec. \ref{subsec:Results_AD}, and for Compressed Replay in Sec. \ref{subsec:results_f1_SCALE}.

\end{itemize}
As for the approaches that can be used as AD Module only, we consider:
\begin{itemize}
    \item \textbf{Inpaint}: We tried to use the same architecture as RIAD \cite{ZAVRTANIK2021107706} in our study but found that the architecture was very sensitive to the continual learning setting, and we were unable to obtain satisfactory results. As an alternative, we adopted the spirit of the RIAD approach (i.e., inpainting) and proposed a different architecture, a Pix2Pix model \cite{isola2017image}, for the inpainting task. In our version, training is done by masking random areas of the images. During reconstruction, the same image is fed into the model multiple times (with different masks each time), and averaging is used if the same pixel is masked multiple times.

\end{itemize}

\subsection{Considered CL Methods}
\label{subsec:considered_cl_strategies}
We evaluate the different architectures in terms of performance for the Memory Module and the AD Module.
Such evaluation is performed considering different CL methods applied to the models.
\begin{itemize}

    \item \textbf{Single Model}: As an upper bound, we train a different model for each task. 

    \item \textbf{Fine-tuning}: As a lower bound, we consider the fine-tuning approach, in which a model is presented sequentially only with data from the current task. 

    \item \textbf{Replay High Mem}:
    In the Replay technique, $n$ images are selected randomly from the replay memory and concatenated with other $n$ images selected randomly from the current task's data. The batch is then given to the neural network for training.
    
    However, in this version of Replay, called Replay High Mem, it is made the ideal assumption to have enough memory to store all samples.
    This means that the memory can capture the original distribution completely.

    This is a very optimistic assumption, and this approach is unfeasible in a real-world scenario.
    This method represents the best performance that the Replay approach could obtain if the memory is not constrained.
    
    The \textit{Replay High Mem} approach serves as an upper bound for other Continual Learning methods like \textit{Replay Low Mem} and \textit{Compressed Replay} (discussed below). These alternatives consider memory and computational constraints, making them more realistic and feasible for Continual Learning. 
    
    \item \textbf{Replay Low Mem}: In this version, the Replay technique assumes a finite memory size to store only a minimum portion of old data. Here, it adopts the same method of \textit{Replay High Mem} but uses a reduced memory, making the implementation of the approach realistic in real-world applications.
    More in detail, we consider a constrained memory size of $n$ images such that: $n << |D|$ where $D$ represents the entire dataset. In the experiments, $n$ is set to $40$ images, representing less than $2\%$ of the total dataset.
    This means that the total memory size is limited to $40$ images, i.e., four images for each class at the end of the training.

    \item \textbf{Compressed Replay}:
    When dealing with Replay, it should be noted that the risk of overfitting is high when only a few samples represent a task.
    Given that memory size can influence final results,  the ability to save more samples has the potential to reduce forgetting because the distribution in memory of an old task is more similar to the original one (assuming that the quality of compressed samples is high enough). 
    So, in Compressed Replay, the images are compressed. 

    The Compressed Replay is tested for the models CAE and VAE (further discussion below). 
    Moreover, we test the Compressed Replay with the SR model (i.e., our novel approach SCALE) using the procedure described in Sec. \ref{Sec:SCALE}
    The Compressed Replay is first tested for each model in terms of Memory Module by assessing their ability to compress and reconstruct the images with the results shown and discussed in Sec. \ref{subsec:Results_FID}.
    Then, we consider the Compressed Replay based on the AD performance, showing the results in Sec. \ref{subsec:results_f1_SCALE} and in Tab. \ref{Tab:summary_table_f1_part2}.
    In some cases, like for the VAE, a single model acts as Memory and AD modules.
    In other cases, a model will act as Memory (e.g., SCALE), and another will act as an AD Module (e.g., Inpaint).

    \item \textbf{Generative Replay}: In the case of Generative Replay \cite{shin2017continual}, we use models such as VAEs to generate images without storing any additional information other than the model itself. This can be very useful because it eliminates the need to save any data in memory. 
    However, in previous experiments, the Generative Replay has a high level of forgetting when the data distribution to learn is very complex, like for the CIFAR10 dataset \cite{lesort2019generative}.
    \end{itemize}

\textbf{How an Autoencoder can perform Compressed Replay?}
    For autoencoder architectures (like VAE and CAE), we consider the same architecture to act as both Memory and AD functions.
    When the AE sees new images, it compresses them using the Encoder and stores them in the latent space.
    During the training for a new task, the compressed samples are retrieved from the memory and decompressed using the Decoder.
    The decompressed images are concatenated with the images of the new task and used to update the AE model (Encoder and Decoder).
    As said before, the ability to reconstruct images can be used not only to compress images and perform Compressed Replay but also to detect anomalies.
    Indeed, the difference between the input and reconstructed images can be used to create anomaly maps and perform Anomaly Detection during inference.
    
    We utilize the same amount of memory (in bytes) for both approaches to achieve a fair comparison between Compressed Replay and Replay. We define the memory size as $C$, which is the number of bytes needed to store $n$ images using the Replay approach. For Compressed Replay, some compressed samples will be stored, taking up a maximum of $C$ bytes of storage.

    We compare the CL approaches based on two abilities: the AD performance and the quality of the reconstructed images. 
    We also consider the compression factor as a key metric for the Compressed Replay approach.

\subsection{Evaluation Metrics for Anomaly Detection}
\label{subsec:metrics_AD}
Several metrics are commonly used to evaluate the performance of Anomaly Detection models on the MVTec AD dataset. The ROC AUC score at the pixel and image levels is the most widely used, but other metrics, such as the f1 score and IoU, are also considered. 
In the Anomaly Detection literature, metrics such as the f1 score are generally believed to be more fair when dealing with imbalanced datasets, which is often the case with Anomaly Detection \cite{saito2015precision}. 
Therefore, in this study, we use the f1 score as the main metric to assess a model's ability to detect anomalies at pixel-level.

\subsection{Evaluation for image reconstruction quality}
\label{subsec:metrics_image_quality}

To evaluate the quality of the reconstruction performed by the models used in the Memory Module, we will use the Fréchet Inception Distance (FID) \cite{heusel2017gans}, which is a commonly used metric for evaluating generative models \cite{lesort2019generative}. 
This type of evaluation is useful for analyzing the performance of all architectures that can function as Memory Modules under various CL methods.
The FID is calculated as:
\begin{equation}
    FID = ||\mu_r - \mu_g||^2+Tr( \Sigma_r+\Sigma_g-2(\Sigma_r\Sigma_g)^\frac{1}{2} )
\end{equation}
The statistics $(\mu_r,\Sigma_r)$ and $(\mu_g,\Sigma_g)$ are the activations of a specific layer of a discriminative neural network trained on ImageNet for real and generated samples, respectively.
A lower FID indicates that real and generated samples are more similar, as measured by the distance between their activation distributions.

\subsection{Continual Learning Metrics}
\label{subsec:metrics_cl}
\label{sec:metrics}
Following the convention of Continual Learning, we will consider the average value at the end of the training and the forgetting.
Let $s_{i,j}$ be the performance f1 of the model on the test set of task $j$ after training the model on task $i$. To measure performance in the CL setting, we introduce the following metrics:
\begin{description}
\item[Average f1] 
The average f1 score $S_T \in [0,1]$ at task T is defined as:
\begin{equation}
    S_T = \frac{1}{T} \sum_{j=1}^T s_{T,j}
\end{equation}

\item[Average Forgetting] $F_T \in [-1,1]$, the average forgetting measure at task T, is defined as:
\begin{equation}
    F_T = \frac{1}{T-1} \sum_{j=1}^{T-1} max_{l \in \{1,\cdots,T-1\} } \frac{s_{l,j}-s_{T,j}}{s_{l,j}}.
\end{equation}
Concerning the original definition used in \cite{chaudhry2018riemannian}, we are scaling with respect to the maximum f1 score, as done in \cite{kim2020imbalanced,pezze2022multi}. This is done to compare the forgetting among tasks with very different scores. Notice that the closer the metric $F_T$ is to 1, the higher the forgetting is.
What was defined above for the metric f1 will be redefined in a very similar way for the metric FID, with the consideration that the goal of FID is to be minimized, while with f1, we want to maximize it.

\end{description}

\section{Experimental results}
\label{Sec:Results}

\begin{figure}[!th]
    \begin{subfigure}[t]{0.49\textwidth}
        \centering
        \includegraphics[width=\textwidth]{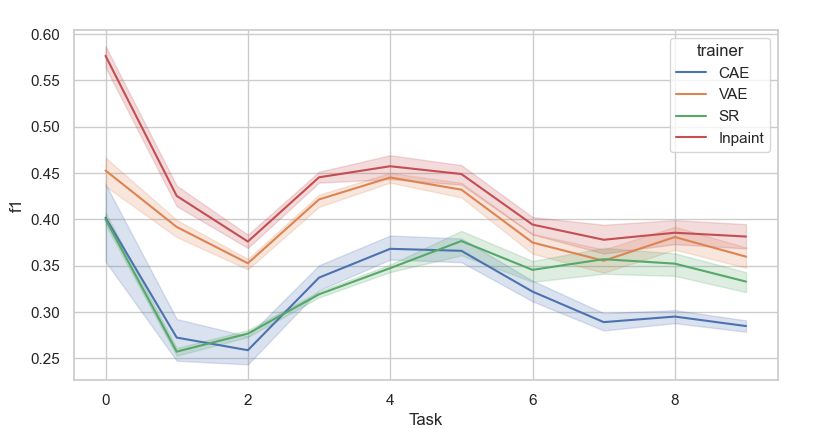}
        \caption{Single Model}
         \label{fig:LinePlotF1_by_strategy_a}
    \end{subfigure}
    \begin{subfigure}[t]{0.49\textwidth}
        \centering
        \includegraphics[width=\textwidth]{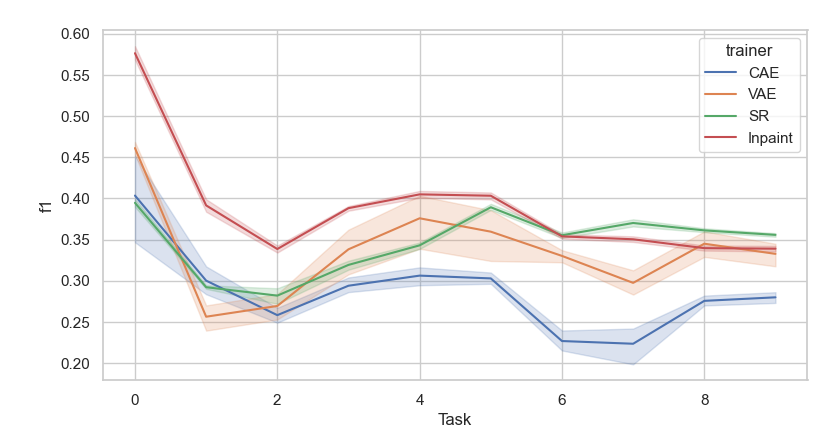}
        \caption{Replay High Mem}
         \label{fig:LinePlotF1_by_strategy_b}
    \end{subfigure}
    \bigskip \\
    \begin{subfigure}[t]{0.49\textwidth}
        \centering
        \includegraphics[width=\textwidth]{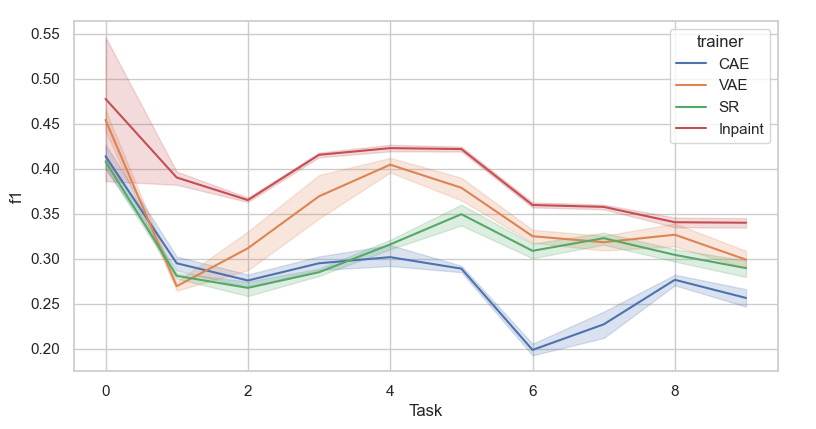}
        \caption{Replay Low Mem}
         \label{fig:LinePlotF1_by_strategy_c}
    \end{subfigure}
    \begin{subfigure}[t]{0.49\textwidth}
        \centering
        \includegraphics[width=\textwidth]{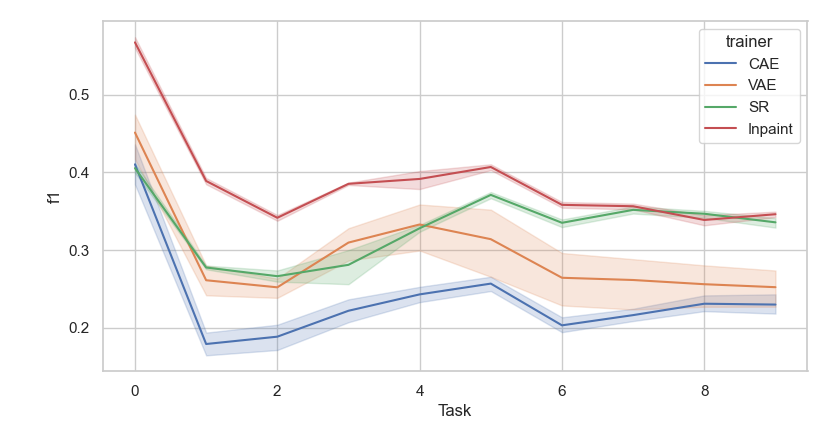}
        \caption{Compressed Replay}
         \label{fig:LinePlotF1_by_strategy_d}
    \end{subfigure}    
    \caption{Each plot shows the performance for a CL method comparing different architectures in terms of AD performance, i.e., Average f1 score.  The Average f1 metric is shown on axis y, and the index of the current training task is shown on axis x.}
    \label{Fig:LinePlotF1_by_strategy}
\end{figure}

In Sec. \ref{subsec:Results_AD}, we discuss the results shown in Tab. \ref{Tab:summary_table_f1_part1} and Fig. \ref{Fig:LinePlotF1_by_strategy}, describing the AD performance of all models with the CL approaches: Single Model, Fine-tuning, Replay High Mem, Replay Low Mem, Generative Replay.
Then, in Sec. \ref{subsec:Results_FID}, we test our novel approach, SCALE (Compressed Replay with SR), and compare it with other techniques like VAE regarding reconstruction image quality.
After assessing the superiority of SCALE in terms of reconstruction image quality, in Sec. \ref{subsec:comparison_AD_SCALE} we are interested in evaluating how well SCALE as a Memory Module is useful for the AD techniques.

\begin{table}[th]
\caption{Summary table for the AD performance under different CL techniques and architectures for the Memory and the AD Module. We report the Average f1 value ($\uparrow$ stands for "the higher the better") and, within round brackets, the forgetting metric. The best combination is in \textbf{bold}. We can observe that Replay Low Mem approach is effective in the CL setting reaching values close to the upper bound Single Model for all the AD techniques.}
\label{Tab:summary_table_f1_part1}
\centering
\normalsize
\begin{tabular*}{\textwidth}{@{\extracolsep{\fill}}lcccc}
\toprule
\textbf{CL Approach \textbackslash $~$ AD Model}          & \textbf{CAE} & \textbf{VAE} & \textbf{SR} & \textbf{Inpaint} \\
\midrule
\textbf{Single Model}      & 0.28 & 0.36 & 0.33 & 0.38 \\
                           &  -  &  -  &  -  &  -  \\
\midrule
\textbf{Fine-Tuning}       & 0.09 & 0.11 & 0.14 & 0.11 \\
                           & (66.35\%) & (68.19\%) & (59.39\%) & (72.73\%) \\
\midrule
\textbf{Replay High Mem}   & 0.28 & 0.33 & \textit{0.36} & 0.34 \\
                           & (13.22\%) & (6.38\%) & (5.72\%) & (11.25\%) \\
\midrule
\textbf{Replay Low Mem}    & 0.26 & 0.30 & 0.29 & \textbf{0.34} \\
                           & (19.88\%) & (21.2\%) & (22.23\%) & (13.27\%) \\
\midrule
\textbf{Generative Replay} & - & 0.11 & - & - \\
                           &  -  & (63.68\%) &  -  &  -  \\
\bottomrule
\end{tabular*}
\end{table}

\FloatBarrier
\subsection{{Anomaly Detection performance}}
\label{subsec:Results_AD}
In this section, we will evaluate the performance of the following AD techniques in the CL setting: CAE, VAE, Inpaint, and SR.
For all these AD models, we evaluate them with several CL methods such as:
Single Model, Fine-Tuning, Replay High Mem, Replay Low Mem, and Generative Replay. 
Fine-tuning is considered the lower bound of performance, and Single Model is the upper bound for the CL methods.
Similarly, also Replay High Mem is considered an upper-bound for all replay-based approaches since it violates the memory constraints and assumes enough memory to memorize all samples of all seen tasks, which is not feasible in a realistic scenario.
This model is used to compare all the CL approaches, such as Replay Low Mem, Compressed Replay, and Generative Replay, which consider a finite memory size.

Table \ref{Tab:summary_table_f1_part1} provides a comprehensive overview of the obtained results, while Fig. \ref{Fig:LinePlotF1_by_strategy} presents the results in greater detail, revealing how performance changes over time when visiting in sequence the tasks. 
As expected, the Fine-Tuning approach, where the model is trained only on the current data, has very low performance for all AD models, e.g., an f1 score of 0.11 for VAE.

In contrast, replay-based approaches like Replay High Mem and Replay Low Mem substantially enhance the performance of all AD models. 
For example, VAE with Replay Low Mem achieves an f1 score of 0.3 which is very close to the 0.36 f1 score of the upper bound Single Model.
We can observe that such values are not so distant from those of the Single Model (upper bound), which has an f1 score of 0.36.
In particular, when comparing the different AD models, we can observe that the best results for Replay Low Mem are obtained by the model Inpaint, which obtains an f1 score of 0.34 and a forgetting value of 13.27\%.

We also tested Generative Replay for the VAE model, which is the only AD model to generate images from random noise, avoiding the costs of memorizing old samples in a replay buffer. However, the forgetting is very high, causing the performance to decrease quickly. The VAE with Generative Replay has an f1 score of 0.11 and a forgetting rate of 63.68\%. 

Therefore, we can affirm that replay-based approaches proved effective in solving AD in the CL setting. 
Specifically, we can assess that the Replay Low Mem is a realistic approach to defeating forgetting and enabling AD techniques in the CL setting.
However, as noted before, the use of replay memory could be a too heavy burden for resource-constrained devices or when the consumed memory is a big constraint.

Therefore, we evaluate different Compressed Replay approaches (including our novel approach, SCALE) in Section \ref{subsec:Results_FID}, focusing on image reconstruction quality and compression factor.

Then, in Sec. \ref{subsec:results_f1_SCALE}, different Memory Modules and AD Modules are evaluated to see if approaches like SCALE are a feasible way to compress the replay memory for the AD problem.

\FloatBarrier
\subsection{Quality for reconstructed images}
\label{subsec:Results_FID}

\begin{table}
\centering
\normalsize
\caption{In this table, we compare the reconstruction quality of the Memory Module. For each model and CL technique, the FID metric (lower is better) is reported, while the forgetting is represented in round brackets. Note that for each CL technique, the best model is SR. In bold the best value among the practical CL approaches. }
\label{Tab:summary_table_fid}
\begin{tabular}{|c|c|c|c|} 
\hline
                           & \multicolumn{3}{c|}{Average FID $\downarrow$}                                                                                                                                                                           \\ 
\hline
\textbf{CL Method}          & \textbf{CAE}                                              & \textbf{VAE}                                                                & \textbf{SR}                                                                   \\ 
\hline
\textbf{Single Model}      & 266.52                                                    & 219.40                                                                      & \textit{125.15}                                                               \\ 
\hline
\textbf{Fine-Tuning}                & \begin{tabular}[c]{@{}c@{}}376.64\\(73.65\%)\end{tabular} & \begin{tabular}[c]{@{}c@{}}399.44\\(91.33\%)\end{tabular}                   & \begin{tabular}[c]{@{}c@{}}\textit{244.25}\\(194.31\%)\textit{}\end{tabular}  \\ 
\hline
\textbf{Replay High Mem}   & \begin{tabular}[c]{@{}c@{}}216.82\\(7.50\%)\end{tabular}  & \begin{tabular}[c]{@{}c@{}}255.84\\(11.93\%)\end{tabular}                   & \begin{tabular}[c]{@{}c@{}}\textit{89.53}\\(7.74\%)\textit{}\end{tabular}     \\ 
\hhline{|====|}
\textbf{Replay Low Mem}    & \begin{tabular}[c]{@{}c@{}}227.88\\(14.70\%)\end{tabular} & \begin{tabular}[c]{@{}c@{}}245.87\\(14.54\%)\end{tabular}                   & \begin{tabular}[c]{@{}c@{}}\textit{111.87}\\(40.6\%)\textit{}\end{tabular}    \\ 
\hline
\textbf{Compressed Replay} & \begin{tabular}[c]{@{}c@{}}248.42\\(20.33\%)\end{tabular} & \begin{tabular}[c]{@{}c@{}}316.48\\(36.17\%)\end{tabular}                   & \begin{tabular}[c]{@{}c@{}}\textbf{105.57}\\\textbf{(14.66\%)}\end{tabular}   \\ 
\hline
\textbf{Generative Replay} & -                                                         & \begin{tabular}[c]{@{}c@{}}\textit{405.34}\\\textit{(77.84\%)}\end{tabular} & -                                                                             \\
\hline
\end{tabular}
\end{table}

\begin{table}[th]
\centering
\normalsize
\begin{tabular}{|c|c|c|}
\hline
\multicolumn{1}{|l|}{\textbf{\begin{tabular}[c]{@{}l@{}}Method for\\ Compressed Replay\end{tabular}}} & \textbf{Compression Factor} & \multicolumn{1}{l|}{\textbf{FID  $\downarrow$ }} \\ \hline
CAE                                                                                                   & x6                          & 227.88                                            \\ \hline
VAE                                                                                                   & \textbf{x196}               & 245.87                                            \\ \hline
Inpaint                                                                                               & x1                          & 181.74                                            \\ \hline
SCALE (ours)                                                                             & x64                         & \textbf{105.57}                                   \\ \hline
\end{tabular}
\caption{For each model is indicated the ability to compress the data and the quality of reconstruction (FID). In bold the best value based on the metric. VAE shows a high compression factor but terrible reconstruction results, while SCALE shows a good compression while achieving satisfactory reconstruction quality. }
\label{Tab:compression_factors}
\end{table}

\begin{figure}[!th]
    \begin{subfigure}[t]{0.49\textwidth}
        \centering
        \includegraphics[width=\textwidth]{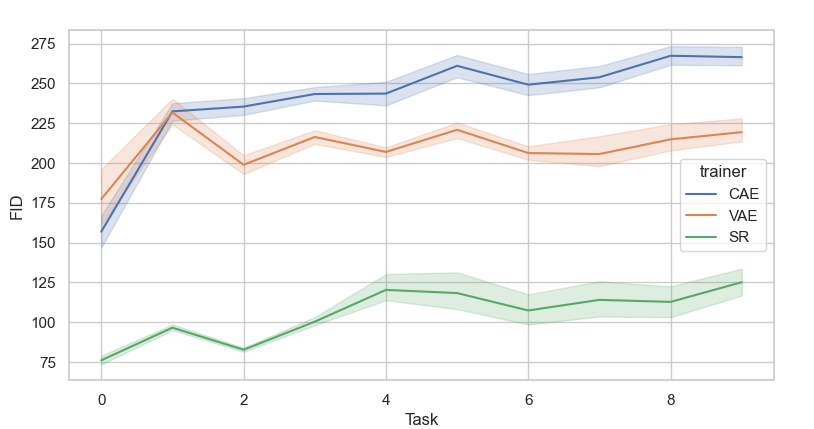}
        \caption{Single Model}
         \label{fig:PlotFID_by_strategy_a}
    \end{subfigure}
    \begin{subfigure}[t]{0.49\textwidth}
        \centering
        \includegraphics[width=\textwidth]{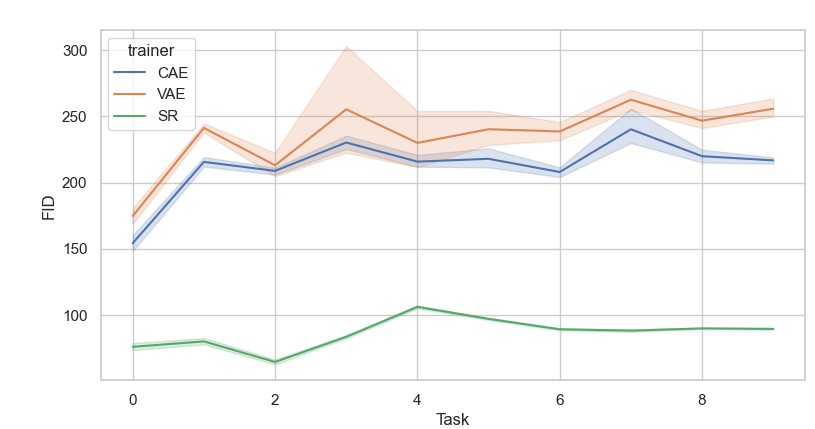}
        \caption{Replay High Mem}
         \label{fig:PlotFID_by_strategy_b}
    \end{subfigure}
    \bigskip \\
    \begin{subfigure}[t]{0.49\textwidth}
        \centering
        \includegraphics[width=\textwidth]{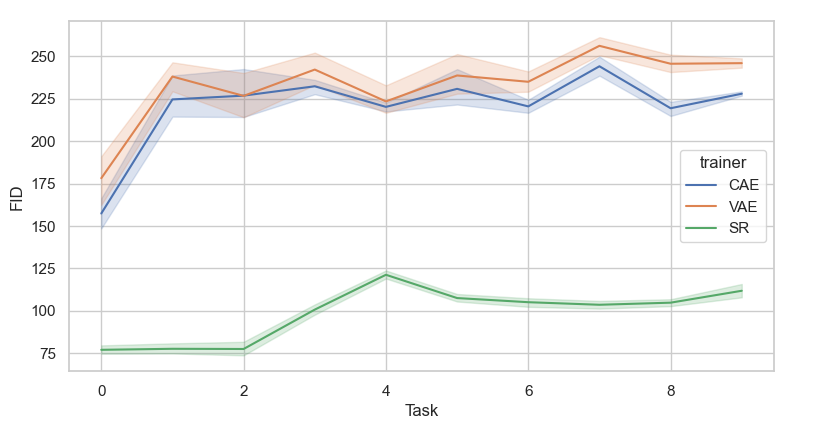}
        \caption{Replay Low Mem}
         \label{fig:PlotFID_by_strategy_c}
    \end{subfigure}
    \begin{subfigure}[t]{0.49\textwidth}
        \centering
        \includegraphics[width=\textwidth]{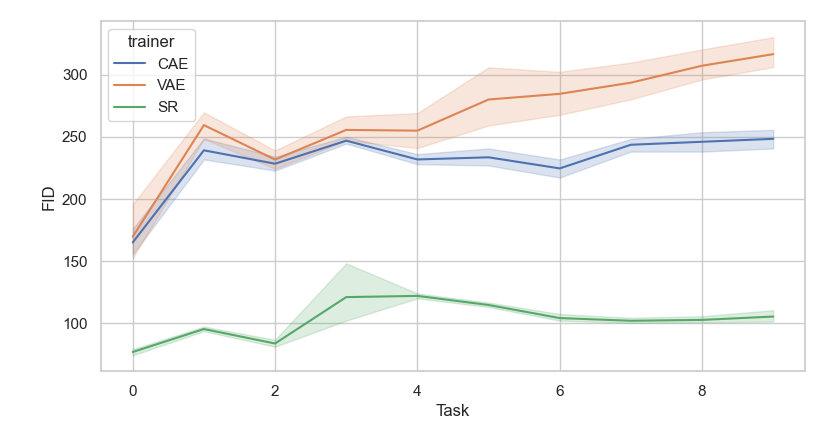}
        \caption{Compressed Replay}
         \label{fig:PlotFID_by_strategy_d}
    \end{subfigure}    
    \caption{The results show the quality of the constructed images for different continual learning methods and AD architectures. Each plot represents a different CL technique, and each line represents a different architecture. The y-axis shows the FID metric, and the x-axis shows the index of the current training task. Each value represents the average FID across all tasks seen so far, with a lower value indicating a higher reconstruction quality. The plots show that the best architecture for each CL method is SR, our approach.   }
    \label{Fig:PlotFID_by_strategy}
\end{figure}

Before comparing Replay Low Mem and Compressed Replay for the AD techniques, we need to evaluate the methods CAE, VAE, and SR in terms of Memory Module.
Our goal is to evaluate the ability of CAE, VAE, and SR to compress and reconstruct the images. 
The Compressed Replay is the CL method that performs this, and the scheme is depicted in Fig. \ref{fig:CLAD_framework_scheme_CompressedReplay}.
Moreover, in Tab. \ref{Tab:summary_table_fid}, we compare Compressed Replay and the results obtained with classic CL approaches like Single Model, Fine-Tuning, Replay High Mem, Replay Low Mem, and Compressed Replay.
In Tab. \ref{Tab:summary_table_fid}, the results are expressed in terms of average FID (where a lower value means a better quality of the reconstructed image) and, in round brackets, the associated percentual forgetting value.
Moreover, it is possible to examine the behavior of each AD model and CL approach in terms of average FID in the plots in Fig. \ref{Fig:PlotFID_by_strategy}. 
Specifically, each plot shows a CL approach while varying the AD model.

In practice, we are interested in results for the Compressed Replay; the other CL approaches are used as upper and lower bounds to understand how much Compressed Replay loses regarding reconstruction quality when considering approaches to reduce replay memory.
The Memory Module aims to provide reconstructed images of old tasks for the AD techniques, as depicted in Fig. \ref{fig:CLAD_framework_scheme_CompressedReplay}.
If the compression factor and the reconstruction quality are satisfactory, we can consider these models as Memory Module. 
The SCALE approach consists of the SR model applied with the Compressed Replay approach with the scheme depicted in Fig. \ref{Fig:framework_SR} and with the procedure described in Sec. \ref{Sec:SCALE}.
The first observation coming from Fig. \ref{Fig:PlotFID_by_strategy} (or Table \ref{Tab:summary_table_fid}) is that the SR model is the best model for each CL method  by a significant margin.
For instance, in the case of Replay High Mem, SR has an FID of 89.53 (lower is better), while the second-best is CAE with 216.82.
In particular, for  Compressed Replay, our approach SCALE (Compressed Replay of the SR model) achieves the best performance compared to the other methods, VAE and CAE.
It can obtain 105.57, while the best second (CAE) has 248.42.
In other words, regardless of the CL technique, the SR model is the best for learning a good representation of the original images.
Therefore, SCALE (Compressed Replay for the SR model using the procedure described in Sec. \ref{Sec:SCALE}) is the optimal choice.
The only model that allows Generative Replay is VAE, but as seen in Table \ref{Tab:summary_table_fid}, the performance degrades very quickly, so it is only shown in the tables and not the plots (the same for Fine-Tuning). 

It should be noted that the compression factors for the CAE, VAE, and SR architectures are 6, 196, and 64, respectively, as denoted in Table \ref{Tab:compression_factors}. Though VAE can obtain a higher compression value, the images produced are significantly lower quality. Indeed, SR obtained a FID of 105.57 (where lower is better), while VAE has 245.87. Moreover, it should be observed that the FID measure is nonlinear; therefore, the difference is even more relevant.
As a final remark, the SR model allows for a good compression factor while maintaining a high image quality in memory.

\FloatBarrier
\subsection{AD performance using Compressed Replay approaches}
\label{subsec:comparison_AD_SCALE}
\label{subsec:results_f1_SCALE}

\begin{table}[th]
\caption{Summary table for the AD performance using the realistic CL methods Replay Low Mem and Compressed Replay. It's important to highlight that when referring to Compressed Replay (\textit{model}), it signifies the incorporation of the \textit{model} as a Memory Module. We report the average f1 score (where higher values indicate better performance) and, within round brackets, the forgetting metric. The best combination is in \textbf{bold}. We can observe that, among the viable replay-based approaches, the best combination is using Inpaint as the AD Module and SCALE(SR for Compressed Replay) as the Memory Module.}
\label{Tab:summary_table_f1_part2}
\centering
\normalsize
\begin{tabular*}{\textwidth}{@{\extracolsep{\fill}}lcccc}
\toprule
\textbf{Memory Module} & \multicolumn{4}{c}{\textbf{AD Module}} \\
\cmidrule{2-5}
& \textbf{CAE} & \textbf{VAE} & \textbf{SR} & \textbf{Inpaint} \\
\midrule
\textbf{Replay Low Mem} & 0.26 & 0.30 & 0.29 & 0.34 \\
                        & (19.88\%) & (21.2\%) & (22.23\%) & (13.27\%) \\
\midrule
\textbf{Compressed Replay} & 0.23 & - & - & 0.25 \\
                   \textbf{(CAE)}              & (25.61\%) &  -  &  -  & (34.00\%) \\
\midrule
\textbf{Compressed Replay} & - & 0.25 & - & 0.26 \\
            \textbf{(VAE)}                     &  -  & (21.87\%) &  -  & (30.27\%) \\
\midrule
\textbf{Compressed Replay} & - & - & 0.34 & \textbf{0.35} \\
     \textbf{(SR)}          &  -  &  -  & (7.99\%) & (10.28\%) \\
\bottomrule
\end{tabular*}
\end{table}

In Section \ref{subsec:Results_AD}, we tested Replay with several implemented AD approaches and showed that it generally performs well in the CL setting. However, as discussed before, the problem with Replay is the limited memory size, which reduces the benefit of this approach.

A solution is to employ compression techniques to obtain compressed representations of the samples.
Therefore, in Section \ref{subsec:Results_FID} we tested different models like VAE to act as Memory Module.
In other words, we tested their capacity when used for Compressed Replay to reconstruct the images well, showing the superiority of our approach SCALE, which is an SR model for Compressed Replay.

In the following part, we discuss the AD performed in conjunction with Memory modules and AD models.
Note that in some cases, the Memory Module and AD Module can coincide like in the cases of CAE and VAE. This is because some models can work as Memory and as AD modules at the same time. 
In other cases, we consider the conjunction of our Memory Module, SCALE, in conjunction with the best AD model, Inpaint.

When examining the results of Tab. \ref{Tab:summary_table_f1_part2}, we can note that the performance of CAE and VAE, which function simultaneously as Memory and AD modules, is lower than using Replay Low Mem directly. This indicates that these models are not sufficiently effective at compressing and decompressing images for use as Memory Modules in the AD problem. Instead, in such cases, replaying the original images, as done in Replay Low Mem, seems to be a better solution.

Instead, when considering the SR model as a Memory and AD module, we can see that its performance is superior to that of the Replay Low Mem.
This is explained by the fact that at parity of memory in bytes, SCALE can save many more samples in memory by compressing them, which allows it to obtain an f1 score of 0.32 compared to 0.29 of Replay Low Mem.

In Sec. \ref{subsec:Results_AD}, we identified Inpaint as the best AD model among those tested. However, when combining the Memory Modules CAE and VAE with Inpaint as the AD model, the performance remained low, with f1 scores of 0.25 and 0.26, respectively, which are lower than the f1 score of 0.34 obtained by Replay Low Mem with Inpaint.

Finally, we evaluated the combination of the best Memory Module identified in Section \ref{subsec:Results_FID}, SCALE, with the best AD model, Inpaint. This combination achieved an f1 score of 0.35, which is higher than the 0.34 f1 score obtained by Replay Low Mem, demonstrating the effectiveness of SCALE in compressing the replay memory.

\FloatBarrier
\section{Conclusions}
\label{Sec:Conclusion}

This research proposes a framework for performing Pixel-Level Anomaly Detection in a setting where the model continually learns new tasks. 
Our framework is structured by a Memory module, which stores information from previous tasks, and an Anomaly Detection module, which identifies anomalies in new data. 
We tested several AD models in the CLAD setting and adapted them to use Replay, proving their effectiveness in remembering how to detect correctly anomalies from previously learned tasks. 

One issue with the Replay is the use of external memory, which could be too heavy for resource-constrained devices.
Therefore, within this framework, we introduce SCALE, which uses a Super Resolution model to compress the samples in memory.  
We show the superiority of SCALE in terms of reconstruction image quality while achieving high compression levels.
Moreover, the conjunction between SCALE as Memory and the best AD model, Inpaint, can achieve optimal performance for Pixel-Level Anomaly Detection.

Future research about the CLAD setting will consider approaches belonging to the embedding similarity-based category, like FastFlow and PatchCore, which weren't considered in this study.

Another interesting research direction would be to test the SCALE approach in other problems different from Anomaly Detection, e.g., classification and Object Detection.
In particular, it would be extremely interesting to see its application in problems such as Semantic Segmentation, where providing samples with precise pixel-level information is extremely relevant, like for the studied Pixel-level Anomaly Detection problem.

\bibliographystyle{elsarticle-num}  
\bibliography{references.bib}

\appendix

\section{Behavior Analysis of Super-Resolution Models in the Continual Learning Setting}
\label{subsec:Result_SR}
\label{subsec:Results_SR}

In the following text, we further study the super resolution problem in the context of Continual Learning. We thoroughly examine how variables such as the number of epochs and how images are saved affect the final performance of the quality of the reconstructed images. This analysis helps us to gain insight into the critical issues facing generative models such as GANs in Continual Learning. Based on this analysis, we conclude that there are at least three critical issues that arise during the training of generative models in CL. 
\begin{enumerate}
    \item The initial quality of the learned distribution is the first critical issue. Based on the number of epochs, we can easily demonstrate that an initial higher quality has a long-term beneficial effect.

    \item 
    \begin{enumerate}
    \item In conventional GANs, images are generated solely by employing a random vector for each new task, resulting in an output image utilized as input for the model in the subsequent task. Nonetheless, this approach introduces perturbations in the final image, and these perturbations accumulate over time, causing a divergence from the original distribution.

    \item Other generative models like VAE and CAE use the compressed representation to reconstruct the image. However, the stored representations of old tasks change each time a new task is trained, causing in this case a perturbation problem as proved in the Tab.     \ref{Tab:scaling_model_comparison_FID} when comparing Compressed Replay and Degenerative Compressed Replay. 
        \end{enumerate}
    
    \item The third critical issue is catastrophic forgetting of the model's weights, which is inherent in all models in the CL setting. Therefore it will not be addressed further in the following discussion.
\end{enumerate}

\begin{figure*}[t]
    \centering\includegraphics[width=0.47\textwidth]{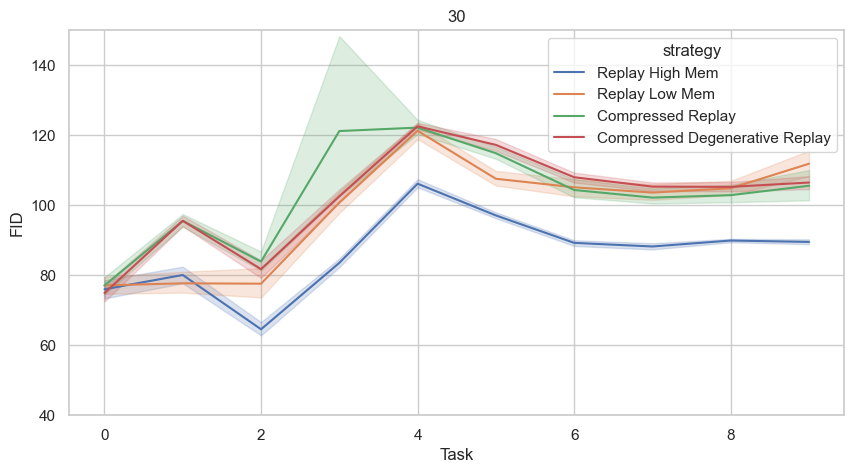}
    \centering\includegraphics[width=0.47\textwidth]{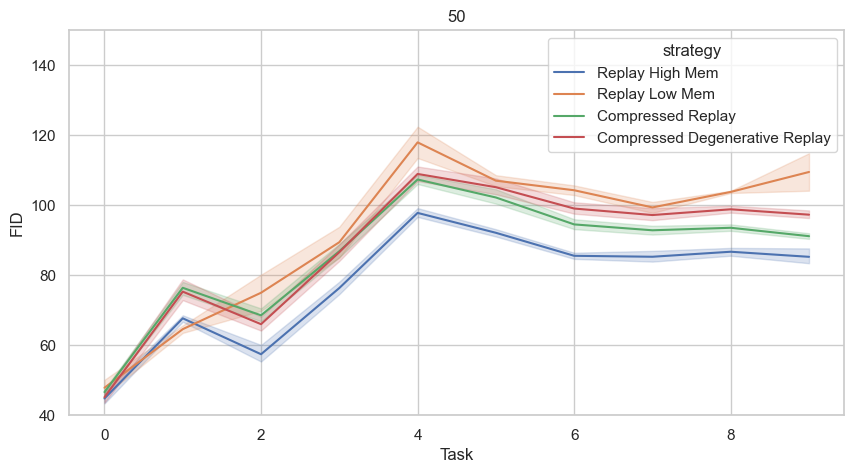}
  \caption{The comparison of the SR model using 30 epochs (left) and 50 epochs (right) with different methods is shown in the figures. The y-axis shows the FID metric, and the x-axis shows the index of the current training task. It can be observed that the Compressed Replay approach performs closer to the Replay High Mem approach (i.e., the optimal performance) when there is an increase in the number of epochs from 30 to 50. }
  \label{Fig:scaling_model_comparison_plot}
\end{figure*}

\textbf{Better initial learned representation reduces forgetting (issue 1)}
\\
Table \ref{Tab:scaling_model_comparison_FID} and Fig. \ref{Fig:scaling_model_comparison_plot} illustrate the impact of varying the number of epochs (30 or 50) on the outcomes of each CL method. Notably, the Replay High Mem method improves FID (lower is better) from 95.2 to 89.5, and the Replay Low Mem method decreases from 112 to 109.5. The most significant enhancement is evident in the Compressed Replay approach, where FID drops substantially from 105.5 to 91. 
\\
Hence, with an increased number of epochs (i.e., an enhanced initial learned representation) the results are, as expected, superior. However, the more intriguing observation is that an improved initial learned representation facilitates a deceleration in the forgetting process. 
This is visible by observing that Compressed Replay with 30 epochs has a forgetting of 14.66\% while Compressed Replay with 50 epochs has a forgetting of 10.88\%.
Therefore, a better initial learned representation allows for better performance over time and reduces the pace of degradation of such performance when visiting new tasks.
Another interesting aspect is that, in terms of percentage increase compared to Replay, Compressed Replay shows a minor decrease at 50 epochs (7\%) compared to 30 epochs (18\%). This suggests that the proposed solution performs very close to the upper bound of Replay High Mem, particularly when the initially learned representation is very good (i.e., a higher number of epochs). 

\begin{table*}
\footnotesize
\centering
\begin{tabular}{|c|c|c|c|}
\hline
\textbf{CL Method}                      & \textbf{epochs} & \textbf{\begin{tabular}[c]{@{}c@{}}Average FID $\downarrow$ \end{tabular} }                                                    & \textbf{\begin{tabular}[c]{@{}c@{}}FID Increase \% \\ wrt Replay High Mem \end{tabular} $\downarrow$ }      \\
\hline
\textbf{\begin{tabular}[c]{@{}c@{}}Single\\Model\end{tabular} }                & 30        & 125.15                                                    & \multicolumn{1}{c|}{-}  \\
\hline
\textbf{\begin{tabular}[c]{@{}c@{}}Replay\\High Mem\end{tabular} }                  & 30        & \begin{tabular}[c]{@{}c@{}}89.53\\(7.74\%)\end{tabular}   & \multicolumn{1}{c|}{-}  \\
\hline
\textbf{Replay Low Mem}                         & 30        & \begin{tabular}[c]{@{}c@{}}111.87\\(40.6\%)\end{tabular}  & 24.95\%                \\
\hline
\textbf{\begin{tabular}[c]{@{}c@{}}Compressed \\ Degenerative \\Replay\end{tabular}} & 30        & \begin{tabular}[c]{@{}c@{}}106.49\\(13.58\%)\end{tabular} & 18.94\%                \\
\hline
\textbf{\begin{tabular}[c]{@{}c@{}}Compressed \\Replay\end{tabular}}             & 30        & \begin{tabular}[c]{@{}c@{}}\textbf{105.57}\\\textbf{(14.66\%)}\end{tabular} & \textbf{17.92\%}               \\
\hhline{|====|}
\textbf{\begin{tabular}[c]{@{}c@{}}Single\\Model\end{tabular} }                 & 50        & 82.06                                                     & \multicolumn{1}{c|}{-}  \\
\hline
\textbf{\begin{tabular}[c]{@{}c@{}}Replay\\High Mem\end{tabular} }                   & 50        & \begin{tabular}[c]{c@{}c@{}}85.27\\(11.18\%)\end{tabular}  & \multicolumn{1}{c|}{-}  \\
\hline
\textbf{Replay Low Mem}                        & 50        & \begin{tabular}[c]{@{}c@{}}109.51\\(59.41\%)\end{tabular} & 28.43\%                \\
\hline
\textbf{\begin{tabular}[c]{@{}c@{}}Compressed \\ Degenerative \\Replay\end{tabular}} & 50        & \begin{tabular}[c]{@{}c@{}}97.31\\(19.08\%)\end{tabular}  & 14.12\%                \\
\hline
\textbf{\begin{tabular}[c]{@{}c@{}}Compressed \\Replay\end{tabular}}              & 50        & \begin{tabular}[c]{@{}c@{}}\textbf{91.17}\\\textbf{(10.88\%)}\end{tabular}  & \textbf{6.92\% } \\
\hline
\end{tabular}
  \caption{Comparison of different CL methods for training an SR model.
 The table shows the results for two different values of the number of epochs (30 and 50). 
 The FID score is shown in the third column, with the percentage of forgetting in round brackets. The fourth column shows the increase in performance with respect to the Replay High Mem, in other words, more a value is lower, and more is close to the optimal. 
 The best results for each number of epochs are highlighted in bold. }
  \label{Tab:scaling_model_comparison_FID}
\end{table*}

\textbf{Why SCALE has better performance than GANs (issue 2.a)}
\\
As described before, the Generative Replay approach for CL uses generative models like GANs to generate images on the fly belonging to old tasks and combine them with the new data to update the model.
However, it has resulted in a high degeneration of reconstructed images, confirmed by the results shown in the article \cite{lesort2019generative} and in Sec. \ref{subsec:Results_AD}.
In our proposed approach called SCALE, we use an SR model; in particular, we consider a Pix2Pix model \cite{isola2017image}, a general-purpose model for image-to-image translation composed of a cGAN.
The results discussed in Sec. \ref{subsec:Results_FID} prove that SCALE can maintain old knowledge even though it is a generative model like the ones used for Generative Replay.
The difference is that Pix2Pix uses a cGAN, which, unlike the classic GAN \cite{mirza2014conditional}, is conditioned not only by random noise but also by an input image.
Therefore, we show in the results of Sec. \ref{subsec:Results_FID} how conditioning to an input image plus the noise can result in a method that can still generate images with a low forgetting even though the output still has a random component.

\textbf{Why SCALE has better performance than models like VAE and CAE (issue 2.b)}
\\
Similarly, other generative models like VAE and CAE use the compressed representation to reconstruct the image. However, the stored representations of old tasks change each time a new task is trained, causing, in this case, a perturbation problem.
This aspect is simulated in SCALE by proposing a method called Degenerative Compressed Replay; contrary to the Compressed replay, this version saves new representations.
\\
The results are shown in Tab \ref{Tab:scaling_model_comparison_FID} for both 30 and 50 epochs.
When comparing the Compressed Replay and Compressed Degenerative Replay, we can see that the positive effect of Compressed Replay in comparison to Compressed Degenerative Replay increases with the number of epochs. 
Compressed Replay exhibits an FID of 105.57 with 30 epochs (91.17 with 50 epochs), whereas Degenerative Compressed Replay registers an FID of 106.49 with 30 epochs (97.31 with 50 epochs). Put differently, there is a 0.92 gap favoring Compressed Replay at 30 epochs and a 6.14 gap at 50 epochs.
\\
However, while Compressed Degenerative Replay is inferior to Compressed Replay, the table shows that Compressed Degenerative Replay with 50 epochs is superior to Compressed Replay with 30 epochs. This means that the main factor influencing performance is the number of epochs (first issue), i.e., the initial quality of the learned representation.

\end{document}